\def\year{2021}\relax
\documentclass[letterpaper]{article} 
\usepackage{aaai21}  
\usepackage{times}  
\usepackage{helvet} 
\usepackage{courier}  
\usepackage[hyphens]{url}  
\usepackage{graphicx} 
\usepackage{natbib}  
\usepackage{caption} 
\usepackage[switch]{lineno} 
\nocopyright
\usepackage{natbib}

\usepackage{listings}
\usepackage{amsmath}
\usepackage{amsfonts}
\usepackage{MnSymbol}

\newtheorem{example}{Example}
\newtheorem{theorem}{Theorem}

\newtheorem{proposition}{Proposition}

\newcommand{\F}{{\cal F}}
\newcommand{\U}{{\cal U}}
\newcommand{\R}{{\cal R}}
\newcommand{\V}{{\cal V}}

\newcommand{\union}{\cup}
\newcommand{\sat}{\models}

\newcommand{\commentout}[1]{}

\newtheorem{definition}{Definition}
\newcommand{\dfn}{\begin{definition}}
\newcommand{\bbox}{\vrule height7pt width4pt depth1pt}
\newcommand{\edfn}{\bbox\end{definition}}
\newcommand{\thm}{\begin{theorem}}
\newcommand{\ethm}{\end{theorem}}
\newcommand{\prf}{\noindent{\bf Proof:} }
\newcommand{\eprf}{\bbox\vspace{0.1in}}
\newcommand{\xam}{\begin{example}}

\title{The Counterfactual NESS Definition of Causation}
\title{The Counterfactual NESS Definition of Causation}
\author {
    Sander Beckers \\
}

\affiliations{
    Munich Center for Mathematical Philosophy \\
    Ludwig Maximilian University, Munich \\
    srekcebrednas@gmail.com
}

\begin{document}
\maketitle

\begin{abstract}
In previous work with Joost Vennekens I proposed a definition of actual causation that is based on certain plausible principles, thereby allowing the debate on causation to shift away from its heavy focus on examples towards a more systematic analysis. This paper contributes to that analysis in two ways. First, I show that our definition is in fact a formalization of Wright's famous NESS definition of causation combined with a counterfactual difference-making condition. This means that our definition integrates two highly influential approaches to causation that are claimed to stand in opposition to each other. Second, I modify our definition to offer a substantial improvement: I weaken the difference-making condition in such a way that it avoids the problematic analysis of cases of preemption. The resulting {\em Counterfactual NESS definition} of causation forms a natural compromise between counterfactual approaches and the NESS approach.
\end{abstract}

\section{Introduction}

\noindent Causal models have become very popular tools to offer definitions of {\em actual causation}, or token causation, in both philosophy and in computer science. Computer scientists (as well as scientists in general) like them because their interventionist semantics fit naturally with the manner in which scientists perform and conceptualize experiments. Philosophers like them because they offer a non-reductive semantics of counterfactuals that avoids the many issues facing a possible-world semantics. It should therefore come as no surprise that the definitions of causation which are expressed using these models fall within the broadly counterfactual approach, which includes interventionist approaches. 

Two such definitions have been developed recently by Vennekens and myself (henceforth the {\em BV definition} \cite{beckers17} and the {\em BV definition-2} \cite{beckers18a}). The proposed definitions by Beckers and Vennekens (BV from now on) distinguish themselves positively from a range of other definitions that are based on that of \cite{halpernpearl05a} -- {\em HP-style definitions} -- by an explicit focus on certain underlying principles that a definition of causation should satisfy, instead of a focus on the -- potentially endless -- debate regarding many examples and the intuitions they invoke. Both definitions are closely related: the BV definition-2 builds on the simpler BV definition by adding temporal considerations.

The first aim of this paper is to show that the BV definition is in fact a formal integration of the influential {\em Necessary Element of a Sufficient Set} definition of causation by \cite{wright85, wright88, wright11} into the more popular counterfactual approach. In particular, the BV definition turns out to be equivalent to stating that the NESS definition holds in the {\em actual} scenario and that the NESS definition does not hold in at least one {\em counterfactual} scenario. In other words, it combines the NESS approach with a counterfactual difference-making condition. What makes this a surprising result, is that the NESS definition belongs to the regularity approach, which is claimed to stand in opposition to the counterfactual character of causal models. A side-by-side analysis of both approaches reveals that this claim is false.

 The second aim of this paper is to modify the BV definitions to overcome two problematic features: they fail to correctly handle standard cases of both {\em early} and {\em late} preemption, unless one invokes probabilistic and temporal considerations. By looking at the precise relations between all definitions here considered I propose a weakening of the difference-making condition that solves this problem, resulting in the  {\em Counterfactual NESS} definition of causation, or {\em CNESS} for short. This analysis shows the CNESS definition to offer a natural compromise between two important approaches to causation.

The paper is structured as follows. The next section introduces the formalism of Structural Equations Modeling that has become widely adopted in the counterfactual approach to causation. Section \ref{sec:ness} offers a detailed comparison of the NESS approach and the most popular counterfactual approaches, and then goes on to present a formalization of the NESS definition using structural equations models. Section \ref{sec:others} discusses the three other attempts at formalizing the NESS definition that have been put forward, and shows each of them to be incorrect. Finally, the formalized version of the NESS definition is used in Section \ref{sec:bv} to introduce the BV definition of causation and discusses how it relates to HP-style definitions, leading the way to the CNESS definition.

\section{Structural Equation Models}\label{sec:causal}

This section reviews the definition of causal models as understood in the structural modeling tradition started by \cite{pearl:book}.
Much of the discussion and notation is taken from \cite{halpernbook} with little change.

\dfn
A signature $\cal S$ is a tuple $(\U,\V,\R)$, where $\U$
is a set of \emph{exogenous} variables, $\V$ is a set 
of \emph{endogenous} variables,
and $\R$ a function that associates with every variable $Y \in  
\U \union \V$ a nonempty set $\R(Y)$ of possible values for $Y$
(i.e., the set of values over which $Y$ {\em ranges}).
If $\vec{X} = (X_1, \ldots, X_n)$, $\R(\vec{X})$ denotes the
crossproduct $\R(X_1) \times \cdots \times \R(X_n)$. For simplicity, I assume here that $\V$ is finite, as is  $\R(Y)$ for every endogenous variable $Y \in \V$. 
\edfn

Exogenous variables represent factors whose causal origins are outside the scope of the causal model, such as background conditions and noise. 
The values of the endogenous variables, on the other hand, are causally determined by other variables within the model.

\dfn
A \emph{causal model} $M$ is a pair $(\cal S,\F)$, 
where $\cal S$ is a signature and
$\F$ defines a  function that associates with each endogenous
variable $X$ a \emph{structural equation} $F_X$ giving the value of
$X$ in terms of the 
values of other endogenous and exogenous variables. Formally, the equation $F_X$ maps $\R(\U \union \V - \{X\})$  to $\R(X)$,
so $F_X$ determines the value of $X$, 
given the values of all the other variables in $\U \union \V$.  
\edfn

Note that there are no functions associated with exogenous variables;
their values are determined outside the model.  We call a setting
$\vec{u} \in \R(\U)$ of values of exogenous variables a \emph{context}.

The value of $X$ may depend on the values of only a few other variables.  
$X$ \emph{depends on $Y$}
if there is some context $\vec{u}$ and a setting of the endogenous variables  
other than $X$ and $Y$ such that 
if the exogenous variables have value $\vec{u}$, then varying the value
of $Y$ in that 
context results in a variation in the value of $X$; that 
is, there is a setting $\vec{z}$ of the endogenous variables other than $X$ and
$Y$ and values $y$ and $y'$ of $Y$ such that $F_X(y,\vec{z},\vec{u}) \ne
F_X(y',\vec{z},\vec{u})$. We then say that $Y$ is a {\em parent} of $X$. $\vec{PA}_{X}$ denotes all parents of $X$, 
which -- for reasons of notational simplicity -- we also take to include $\U$.

In this paper we restrict attention to \emph{strongly recursive} (or
\emph{acyclic}) models, that is, models where, there is a partial order
$\preceq$ on variables such that if $Y$ depends on $X$, then $X \prec Y$. 
In a strongly recursive model, given a context $\vec{u}$, 
the values of all the remaining variables are determined (we can just
solve for the value of the variables in the order given
by $\preceq$). 
\commentout{
A model is \emph{strongly} recursive if the
partial order $\preceq_{\vec{u}}$ is independent of $\vec{u}$; that is,
there is a partial order $\preceq$ such that $\preceq
= \preceq_{\vec{u}}$ for all 
contexts $\vec{u}$.
}
In a strongly recursive model, we often write the
equation for an endogenous variable as $X = f(\vec{Y})$; this denotes
that the value of $X$ depends only on the values of the variables in
$\vec{Y}$, and the connection is given by the function $f$. For example, we might
have $X = Y + 5$. 
 
An \emph{intervention} has the form $\vec{X} \gets \vec{x}$, where $\vec{X}$ 
is a set of endogenous variables.  Intuitively, this means that the
values of the variables in $\vec{X}$ are set to the values $\vec{x}$.  
The structural equations define what happens in the presence of 
interventions.  Setting the value of some variables $\vec{X}$ to
$\vec{x}$ in a causal 
model $M = (\cal S,\F)$ results in a new causal model, denoted $M_{\vec{X}
\gets \vec{x}}$, which is identical to $M$, except that $\F$ is
replaced by $\F^{\vec{X} \gets \vec{x}}$: for each variable $Y \notin
  \vec{X}$, $F^{\vec{X} \gets \vec{x}}_Y = F_Y$ (i.e., the equation
  for $Y$ is unchanged), while for
each $X'$ in $\vec{X}$, the equation $F_{X'}$ for $X'$ is replaced by $X' = x'$
(where $x'$ is the value in $\vec{x}$ corresponding to $X'$).

Given a signature $\cal S = (\U,\V,\R)$, an \emph{atomic formula} is a
formula of the form $X = x$, for  $X \in \V$ and $x \in \R(X)$.  
A {\em causal formula (over $\cal S$)\/} is one of the form
$[Y_1 \gets y_1, \ldots, Y_k \gets y_k] \phi$, where
\begin{itemize}
\item $\phi$ is a Boolean combination of atomic formulas,
\item $Y_1, \ldots, Y_k$ are distinct variables in $\V$, and
\item $y_i \in \R(Y_i)$ for each $1 \leq i \leq k$.
\end{itemize}
Such a formula is abbreviated as $[\vec{Y} \gets \vec{y}]\phi$. The special case where $k=0$ is abbreviated as $\phi$.
Intuitively, $[Y_1 \gets y_1, \ldots, Y_k \gets y_k] \phi$ says that $\phi$ would hold if $Y_i$ were set to $y_i$, for $i = 1,\ldots,k$.

A causal formula $\psi$ is true or false in a \emph{causal setting}, which is a causal model given a
context. As usual, we write $(M,\vec{u}) \sat \psi$  if the causal
formula $\psi$ is true in the causal setting $(M,\vec{u})$.
The $\sat$ relation is defined inductively. 
$(M,\vec{u}) \sat X = x$ if the variable $X$ has value $x$
in the unique (since we are dealing with recursive models) solution
to the equations in $M$ in context $\vec{u}$ (i.e., the unique vector
of values that simultaneously satisfies all
equations in $M$ with the variables in $\U$ set to $\vec{u}$).
The truth of conjunctions and negations is defined in the standard way.
Finally, $(M,\vec{u}) \sat [\vec{Y} \gets \vec{y}]\phi$ if 
$(M_{\vec{Y} \gets \vec{y}},\vec{u}) \sat \phi$.

\section{The NESS Definition of Causation}\label{sec:ness} 

Given the vital role of causation for assessing liability and guilt, a proper definition of causation is crucial for the legal domain. It has long been acknowledged that the \emph{sine qua non} (`but for') account of causation, which is still officially endorsed in many legal texts, is painfully inadequate as a general definition of causation. In the context of structural equations, this flawed account can be described as equating causation with {\em counterfactual dependence}.

\dfn\label{def:dependence} Let $(M,\vec{u})$ be a causal setting, let $C$ and $E$ be endogenous variables, and let $c$ and $e$ be values in $\R(C)$ and $\R(E)$ respectively. We say that $E=e$ is {\em counterfactually dependent} on $C=c$ if $(M,\vec{u}) \sat C=c \land E=e$ and there exists a $c' \in \R(C)$ such that $(M,\vec{u}) \sat [C \gets c'] \lnot (E=e)$.\label{def:cd2}
\edfn

\citet{wright85, wright88} has suggested the NESS definition as an alternative definition of causation, in order to deal with the inadequacies of the sine qua non account. (The NESS definition is very similar to Mackie's well-known {\em INUS} condition. Both are based on the work of \cite{harthonore}.) Over the past few decades this definition has been hotly contested in the literature on causation, culminating in Wright's detailed defense of his definition to the most prominent objections and a discussion on how it differs from the INUS condition
\citep{wright11}. Central to this defense is the distinction between \emph{lawful sufficiency} and \emph{causal sufficiency}: the former expresses that some effect $E=e$ can be derived from some condition $\phi$ by a sequence of logical implications, whereas the latter means that $E=e$ can be derived from $\phi$ by a sequence of instantiations of \emph{causal laws}. 

Parallel to the developments in the philosophy of law, Wright's definition also influenced the literature on causation within the formal causal modeling tradition that started with the work of \cite{pearl:book}. Pearl acknowledges the intuitive appeal of the NESS definition, but criticizes it for lacking precisely those features that Wright appeals to in its defense \cite[p. 314]{pearl:book2}: ``This basic intuition [the NESS intuition] is shared by researchers from many disciplines. ... However, all these proposals suffer from a basic flaw: the language of logical necessity and sufficiency is inadequate for explicating these intuitions''. Instead, Pearl proposes the language of structural equations, whose semantics intend to capture \emph{causal}, rather than logical, necessity and sufficiency. In other words, Wright and Pearl seem to be in agreement on two fundamental issues, namely that causal sufficiency should be distinguished from logical sufficiency, and that the NESS definition of causation is in the right spirit. 

Yet as will become clear below, Pearl's own proposal for defining causation using counterfactuals forms a substantial departure of the NESS definition, which does not rely on counterfactuals (or at least not to the same extent, more on this later). As a result of this, Wright is ``fundamentally opposed to counterfactual analyses of causation and skeptical of attempts by Pearl and others to build formalist accounts based on structural equations.''\footnote{Personal communication.} I believe this is a classic case of throwing away the baby with the bathwater: one can perfectly well use structural equations to correctly formalize definitions of causation that do not belong to the counterfactual approach, such as the NESS definition. In order to show this we take a slight detour to look at existing definitions of causation within the structural equations framework.

\subsection{HP-style Definitions of Causation}\label{sec:hpstyle}

The structural equations framework has become a popular formalism for defining actual causation. The definition proposed by \citet{halpernpearl05a} -- HP, from now on -- has been by far the most influential of the lot.
\commentout{
\footnote{It would be more correct to speak of a family of definitions rather than a single one, since various versions have been developed by Halpern and Pearl, starting with the one proposed by \cite{pearl:book}. \cite{halpernbook} gives a detailed overview of the different versions. However, the one cited is usually taken to be the standard version.}
}
\footnote{Various versions have been developed by Halpern and Pearl, starting with the one proposed by \cite{pearl:book}. \cite{halpernbook} gives a detailed overview of the different versions. However, the one cited is usually taken to be the standard version.}  In fact, many of the other definitions using this formalism can be adequately described as offering modifications of the HP definition, and therefore I will refer to them as \emph{HP-style definitions}.\footnote{Too many of such definitions exist to give an exhaustive list, but some of the more well-known are \citep{hitchcock01,hitchcock07,woodward,hall07,weslake,halpernbook}.}

These definitions are often presented with little systematic motivation, focussing instead on capturing the ``right'' intuition for a myriad of examples. Nevertheless, the reader may verify that the following four assumptions can be used to characterize HP-style definitions. 

\begin{enumerate}
\item
{\bf Formalism} As is clear from the above, they assume that causation can be accurately defined using the structural equations framework. This assumption is rather minimal, but for clarificatory purposes it will turn out to be useful to separate it from three further assumptions that HP-style definitions have in common. 

The second and third assumptions characterize the counterfactual tradition of causation started by \cite{lewis73}, which holds that the notions of counterfactual dependence and causation are tightly intertwined:

\item
 {\bf Dependence} Counterfactual dependence  is {\em sufficient} for causation (but not necessary): if $E=e$ is counterfactually dependent on $C=c$ then $C=c$ causes $E=e$. 
\commentout{\footnote{\cite{beckers17} reserve this name for just the assumption that counterfactual dependence is sufficient. However, since there is such widespread agreement that counterfactual dependence is not necessary, there isn't much use in considering both assumptions separately.}
}

\item
 {\bf Counterfactual} There is also a {\em necessary} condition for causation that is counterfactual in nature, so that causation always includes a statement about what would have happened had the cause not occurred. This condition can be made precise using Definition \ref{contribute1} introduced later on, for now the following informal version suffices: if $C=c$ causes $E=e$ in some actual scenario then there exists a $c' \neq c$ such that $C=c'$ is not causally sufficient for $E=e$ in the corresponding counterfactual scenario. 

\item
 {\bf Interventionism} They all share the assumption that the relation between counterfactual dependence and causation (roughly) takes on the following form: $C=c$ causes $E=e$ iff $E=e$ is counterfactually dependent on $C=c$ given an intervention $\vec{X} \gets \vec{x}$ that satisfies some conditions $P$. The divergence between these definitions is to be found in the conditions $P$ that should be satisfied.\footnote{\cite{weslake} offers a detailed analysis of several of these definitions by taking this assumption as his starting point.} 
\end{enumerate}

I suspect that Wright's scepticism towards using structural equations is based on the observation that these assumptions so often go hand in hand. As will become clear, the BV definition and the CNESS definition show that this need not be the case: both satisfy all of {\bf Formalism}, {\bf Dependence}, and {\bf Counterfactual}, but both reject {\bf Interventionism}, and are thus not HP-style definitions.  Therefore opposition to HP-style definitions need not imply opposition to the structural equations framework as such. The family of HP-style definitions form only one particular group of definitions that one can construct using structural equations. 

Wright clearly rejects {\bf Counterfactual} and {\bf Interventionism}, but there is nothing in his writings which suggest that he rejects {\bf Dependence}.
\commentout{\footnote{His rejection of {\bf Counterfactual} becomes clear when discussing the difference between his NESS account and that of Hart and Honor\'e, as he states that the latter ``interpreted the analysis of necessity as a hypothetical counterfactual analysis rather than a real world factual analysis'' \cite[p. 287]{wright11}. On the same page, he says:
\begin{quote} 
Although I initially referred to the analysis of necessity in both the NESS account and the sine qua non account as a counterfactual analysis, ... , I have always insisted that the analysis is (or should be) a real-world `covering law' matching of actual conditions against the required elements of the relevant causal generalisations rather than a counterfactual `possible worlds' exploration of what might have occurred in the absence of the condition at issue. ... The existence of these two very different methods of analysing necessity is noted by Moore, who however incorrectly describes both methods as `counterfactual' and focuses on the second approach.
\end{quote}
}
}
In fact, as will be argued below, a correct formalization of the NESS definition implies {\bf Dependence} (and falsifies both {\bf Counterfactual} and {\bf Interventionism}). However, in order to get there, it first has to be argued that the NESS definition doesn't conflict with {\bf Formalism}. 

The problem here is that the semantics of structural equations are usually given explicitly in terms of counterfactuals, whereas Wright's notion of causal sufficiency is supposedly distinct from a counterfactual interpretation. Rather than settling this issue by getting into the notoriously overloaded concept of a counterfactual, it is more fruitful to simply compare side-by-side the two notions that form the basic building blocks of both approaches.
\commentout{\footnote{The quotation in the previous footnote attests to the confusion that arises whenever the term counterfactual is invoked. It is also noteworthy that Wright explicitly connects counterfactuals to possible worlds, for causal models do not require a possible world semantics at all.}
}

\subsection{Structural Equations vs Causal Laws}

The structural equations framework as it was developed by \citet{pearl:book2} is not intended to give a reductive account of causation. Its fundamental constituents, the equations, encode basic and autonomous mechanisms that are assumed to be causal themselves. Concretely, these equations encode {\em scientific laws} \cite[p. 27]{pearl:book2}:
\begin{quote}
The interpretation of the functional relationship in $Y=f_Y(\vec{X})$ is the standard interpretation that functions carry in physics and the natural sciences; it is a recipe, a strategy, or a {\em law} specifying what value nature would assign to $Y$ in response to every possible value combination that $\vec{X}$ might take on.\footnote{I've slightly changed the mathematical notation to make it consistent with the notation used in this paper.} [emphasis in original]
\end{quote}

This interpretation of a structural equation as a scientific law is entirely analogous to Wright's concept of a \emph{causal law}, which he invokes to explain the distinction between lawful and causal sufficiency \cite[p. 289]{wright11}:
\begin{quote}
A causal law is an empirically derived statement that describes a successional relation between a set of abstract conditions (properties or features of possible events and states of affairs in our real world) that constitute the antecedent and one or more specified conditions of a distinct abstract event or state of affairs that constitute the consequent such that, regardless of the state of any other conditions, the instantiation of all the conditions in the antecedent entails the immediate instantiation of the consequent, which would not be entailed if less than all of the conditions in the antecedent were instantiated.
\end{quote}

Moreover, just as a structural equation, Wright's notion of a causal law has a built-in directionality (Ibidem):
\begin{quote}
Another critical feature of causal laws -- and the related concept of causal sufficiency as distinct from mere lawful sufficiency -- is their successional or directional nature, according to which the instantiation of the conditions in the antecedent of the causal law causes the instantiation of the consequent, but not vice versa.
\end{quote}

In light of this, there is no a priori reason why the NESS definition could not be formalized using structural equations. We simply need to interpret structural equations as causal laws.

\subsection{Formalizing the NESS Definition}\label{sec:formalness}

We now state Wright's NESS definition (Ibidem):
\begin{quote}
According to the NESS account as initially elaborated, a condition $c$ was a cause of a consequence $e$ if and only if it was necessary for the sufficiency of a set of existing antecedent conditions that was sufficient for the occurrence of $e$. The required sense of sufficiency, which ... I call `causal sufficiency' to distinguish it from mere lawful strong sufficiency, is the instantiation of all the conditions in the antecedent (`if' part) of a causal law, the consequent (`then' part) of which is instantiated by the consequence at issue. 
\end{quote}

As a first step, we formalize what it means for a set of conditions to be sufficient for the occurrence of a consequent condition in a causal law. For simplicity, we restrict ourselves to consequent conditions that take the form of an atomic formula $E=e$. In a causal model $M$, the equation $F_E$ contains the combination of all causal laws that together determine the value of $E$.\footnote{Obviously this means that a structural equation is only an approximate expression of the causal laws, since it only mentions a small subset of the multitude of conditions that a complete specification of the causal laws would require. This is again entirely in line with Wright's conception of causal laws (Ibid, p.290).}
\commentout{This is again entirely in line with Wright's conception of causal laws (Ibid, p.290): 
\begin{quote}
Our knowledge of causal laws generally is incomplete, and even when it is complete we rarely refer to completely specified causal laws, since such complete specification would be extremely burdensome and unnecessarily detailed and lengthy. We rather employ causal generalisations, which refer to only some of the antecedent conditions in the relevant causal laws and have only as much specificity as is possible and needed in the particular situation.
\end{quote}
}
 For example, the equation $E = A \lor B$ states that there are precisely two causal laws which can produce $E=1$, namely the causal law `if $A=1$ then $E=1$', and the causal law `if $B=1$ then $E=1$'. (As explained earlier, the implication here should not be understood as logical implication.) Therefore ``the instantiation of all the conditions in the antecedent (`if' part) of a causal law the consequent (`then' part) of which is instantiated by the consequence at issue'', is translated as: the parent variables $\vec{PA}_E$ of $E$ take on values $\vec{pa}_E$ such that $f_E(\vec{pa}_E)=e$. 

\dfn\label{contribute1}
Given $\vec{X} \subseteq \V$, we say that $\vec{X}=\vec{x}$ is {\em sufficient} for $E=e$ w.r.t.~$(M,\vec{u})$ if $f_{E}(\vec{x},\vec{u})=e$. (Where $f_{E}(\vec{x},\vec{u})=e$ means that for all settings $\vec{v}' \in \V - \{E\}$ such that the restriction of $\vec{v}'$ to $\vec{X}$ is $\vec{x}$, it holds that $f_{E}(\vec{v}',\vec{u})=e$.)
\edfn

The following straightforward proposition gives an alternative formulation of sufficiency in terms of interventions.

\begin{proposition}
$\vec{X}=\vec{x}$ is sufficient for $E=e$ w.r.t.~$(M,\vec{u})$ iff for all values $\vec{y} \in \R(\vec{Y})$ where $\vec{Y} = \V - (\vec{X} \cup \{ E \})$, it holds that $(M,\vec{u}) \sat [\vec{X} \gets \vec{x}, \vec{Y} \gets \vec{y}] E=e$. (Where we assume that $E \not \in \vec{X}$.)
\end{proposition}

Note that Wright refers to the definition quoted above as the initial version. He offers his preferred version later on, in which the necessity of the cause is made redundant by requiring that the antecedent part of a causal law only states those conditions which are necessary. (Thereby ensuring that any condition $c$ which is part of the antecedent is automatically necessary.) Since structural equations may combine several causal laws into a single equation, I prefer to stick with the first formulation. 

Therefore as a second step, we formalize the necessity of the cause ``for the sufficiency of a set of existing antecedent conditions that was sufficient for the occurrence of $e$''. That the antecedent conditions exist, simply means that in the actual setting under consideration, those conditions were satisfied. That the cause $C=c$ was necessary for the set to be sufficient, means that the set without the cause is no longer sufficient.

\dfn\label{nesscause1} [Direct NESS] $C=c$ {\em directly NESS-causes} $E=e$ w.r.t.~$(M,\vec{u})$ if there exists a $\vec{W}=\vec{w}$ so that the following conditions hold:
\begin{itemize}
	\item $(M,\vec{u}) \sat C=c \land \vec{W} = \vec{w}$;
	\item $\{C=c,\vec{W} = \vec{w}\}$ is sufficient for $E=e$ w.r.t.~$(M,\vec{u})$;
	\item $\vec{W} = \vec{w}$ is not sufficient for $E=e$ w.r.t.~$(M,\vec{u})$.
\end{itemize}
We call $\vec{W}=\vec{w}$ a {\em witness}.
\edfn

The following straightforward result is useful to clarify what makes this a case of {\em direct} causation.

\begin{proposition}\label{pro:par}
If $C=c$ directly NESS-causes $E=e$ w.r.t.~$(M,\vec{u})$ then there exists a witness $\vec{W}=\vec{w}$ for this such that $\vec{W} \subseteq \vec{PA}_{E,\vec{u}}$.
\end{proposition}

\citet{wright11} stresses that the NESS definition is distinct from a counterfactual approach because it expresses {\em weak necessity}, as he calls it, as opposed to {\em strong necessity}, as he calls counterfactual dependence. $C=c$ is weakly necessary for $E=e$ if there is {\em a} sufficient condition such that $C=c$ is necessary for its sufficiency, regardless of what other sufficient conditions there might be. A simple application of Definition \ref{nesscause1} illustrates this distinction. Say the equation for $E$ is $E= (C \land B) \lor (\lnot C \land A)$, and we are considering a context such that $A=1$, $B=1$, and $C=1$. Then $C=1$ is a direct NESS-cause of $E=1$, because it is a necessary element of the sufficient set $\{C=1,B=1\}$. The fact that in the counterfactual scenario $C = 0$ becomes sufficient for a different set that would make $E=e$ true -- namely $\{C=0, A=1\}$ -- is entirely irrelevant. Yet it is the existence of that set which explains why $E=1$ is not counterfactually dependent on $C=1$.
\commentout{\footnote{Strevens offers a similar comment to explain how Mackie's interpretation of difference-making differs from the standard counterfactual one \citep[p. 7]{strevens07}: 

\begin{quote}
let me point out two salient differences between Mackie's difference-making and difference-making as defined using natural language counterfactuals. ... 
second, that on the natural language account, you try to remove $c$ only from a single ``sufficient condition'' for $e$, namely, the state of the entire world at the appropriate time, whereas on Mackie's account, you may try to remove $c$ from any number of different sufficient conditions (and there will always be many such conditions). The putative cause $c$ need only be essential to one of these sufficient conditions in order to qualify as a cause.
\end{quote}
}
}

Although we have now formalized the NESS definition as it is stated explicitly in the above quote, there is a crucial part which is made explicit only in the second version: causation is the transitive closure of the relation we have just defined, i.e., we should also consider a sequence of atomic formulas that satisfy the previous definition.
\commentout{.\footnote{For completeness I state the second version in full \citep[p. 291]{wright11}: 
\begin{quote}
an actual condition $c$ was a cause of an actual condition $e$ if and only if $c$ was a part of (rather then [sic] being necessary for) the instantiation of one of the abstract conditions in the completely instantiated antecedent of a causal law, the consequent of which was instantiated by $e$ immediately after the complete instantiation of its antecedent, or {\em (as is more often the case) if $c$ is connected to $e$ through a sequence of such instantiations of causal laws}. [emphasis mine]
\end{quote}
} 
}

\dfn\label{nesscause2} [NESS] $C=c$ {\em NESS-causes} $E=e$ w.r.t.~$(M,\vec{u})$ if there exists a chain of direct NESS causes from $C=c$ to $E=e$. (I.e., there exist $C_1=c_1$, $\ldots$, $C_n=c_n$, so that $C=c$ is a direct NESS cause of $C_1=c_1$, $\ldots$, and $C_n=c_n$ is a direct NESS cause of $E=e$.)
\edfn

\commentout{
I now present two paradigmatic examples from the literature to illustrate how the NESS definition works.

\subsection{Examples}

First we present a nuanced case of overdetermination: there are several individual conditions, the combination of a large number of which is sufficient for some effect. The nuance lies in the fact that unlike in cases of symmetric overdetermination, none of the conditions are individually sufficient. The NESS definition correctly identifies each such individual condition $X=x$ to be a cause, by focussing on sets of conditions such that $X=x$ becomes necessary for securing the outcome.

Say we are considering a simple majority-based voting example, involving three voters. Assume that all three of them vote in favor of the bill. The following equation formalizes this situation: $Bill = (Vote_1 \land Vote_2) \lor (Vote_1 \land Vote_3) \lor (Vote_2 \land Vote_3)$. Clearly, for any choice $i,j$ with $i \neq j$, the set $\{Vote_i=1,Vote_j=1\}$ is minimally sufficient for $Bill=1$, and thus each vote is a cause of the bill passing.

Second we present a standard case of preemption by  \cite[p. 276]{hitchcock01}. Preemption cases form the most well-known challenge for any account of causation. Such a case consists of two causal processes that are in competition to produce some effect, and only one of them succeeds. 
\begin{example} [Backup]\label{ex:backup}
An assassin-in-training is on his first mission. Trainee is an excellent shot: if he shoots his gun, the bullet will fell Victim. Supervisor is also present, in case Trainee has a last minute loss of nerve (a common affliction among student assassins) and fails to pull the trigger. If Trainee does not shoot, Supervisor will shoot Victim herself. In fact, Trainee performs admirably, firing his gun and killing Victim.
\end{example}

It is intuitively clear that Trainee's shot causes Victim to die. The following equations, with the obvious interpretation of the binary variables, form the standard formalization of this story: $Victim = Trainee \lor Supervisor$ and $Supervisor = \lnot Trainee$. The context is such that $Trainee=1$. (Throughout we follow the standard practice of leaving out the equations for endogenous variables such as $Trainee$ that are determined directly by the context.)

To see that $Trainee=1$ is a NESS-cause of $Victim=1$, note that $Trainee=1$ (or $\{Trainee=1\}$, to be entirely correct) is sufficient for $Victim=1$, whereas $\emptyset$ is not. Therefore $Trainee=1$ directly NESS-causes $Victim=1$, and a fortiori also NESS-causes $Victim=1$. 
}

I now present a paradigmatic case of early preemption by  \cite[p. 276]{hitchcock01} to illustrate how the NESS definition works. Preemption cases form the most well-known challenge for any account of causation. Such a case consists of two causal processes that are in competition to produce some effect, and only one of them succeeds. 
\begin{example} [Backup]\label{ex:backup}
An assassin-in-training is on his first mission. Trainee is an excellent shot: if he shoots his gun, the bullet will fell Victim. Supervisor is also present, in case Trainee has a last minute loss of nerve (a common affliction among student assassins) and fails to pull the trigger. If Trainee does not shoot, Supervisor will shoot Victim herself. In fact, Trainee performs admirably, firing his gun and killing Victim.
\end{example}

It is intuitively clear that Trainee's shot causes Victim to die. The following equations, with the obvious interpretation of the binary variables, form the standard formalization of this story: $Victim = Trainee \lor Supervisor$ and $Supervisor = \lnot Trainee$. The context is such that $Trainee=1$. (Throughout we follow the standard practice of leaving out the equations for endogenous variables such as $Trainee$ that are determined directly by the context.)

To see that $Trainee=1$ is a NESS-cause of $Victim=1$, note that $Trainee=1$ (or $\{Trainee=1\}$, to be entirely correct) is sufficient for $Victim=1$, whereas $\emptyset$ is not. Therefore $Trainee=1$ directly NESS-causes $Victim=1$, and a fortiori also NESS-causes $Victim=1$.

\section{Comparison to Other Attempts}\label{sec:others}

As mentioned before, the NESS definition has also been quite influential in the literature on formal approaches to causation. Indeed, this is not the first attempt at formalizing the NESS definition. To the best of my knowledge, three previous attempts have been made, to which we now turn. 

\subsection{Bochman}

\citet{bochman} offers the most recent attempt at formalizing the NESS definition. In fact, he was the first to suggest there is a connection to the work of \cite{beckers18a}.\footnote{It was reading that remark in his paper that provided me with the impetus for writing this one.} However, there are two major problems with his own attempt that conclusively show it to be incorrect: contrary to the NESS definition, his definition of causation is not transitive, and it does not satisfy {\bf Dependence}.\footnote{Concretely, Bochman writes: ``In our theory, general causal inference is transitive, while actual causation is not.'' Also, his Example 8 is a case of counterfactual dependency without causation. \citep[p. 1735]{bochman}}

\subsection{Halpern}

\commentout{
\citet{halpern08} is the latest to offer an attempt at formalizing the NESS definition in the structural equations framework. Unfortunately, just as the attempt discussed below, it was developed before Wright's substantial clarifications in his 2011 article, and therefore it shouldn't come as a surprise that both of them miss the mark. There are three problems with his attempt, only one of which is conclusive evidence that his definition is not a correct formalization of the NESS test. Rather than presenting his definition, I simply state the problems, and leave it to the reader to verify that these apply.

The first problem is nothing but a small technical error: according to his definition, everything that happens causes itself. However, the definition can easily be modified to get rid of this property. The second problem is only a potential problem, which requires further investigation to be settled: it is not clear whether his definition is transitive. If it is not, then just as with Bochman's definition, it conclusively fails as a formalization of the NESS test. Although I was unable to find any violations of transitivity, Halpern does not mention transitivity and nothing in his formulation suggests that he intended his definition to be transitive.\footnote{In personal communication he has confirmed that he did not intend his definition to be transitive and has not looked into whether it is.} The first two problems do not allow us to reach a conclusive verdict, but the third problem does:  Halpern's definition violates {\bf Dependence} (and the NESS definition does not, see Section \ref{sec:bv}.) Consider again Example \ref{ex:backup}, but looking at the context in which $Trainee=0$, and thus it is Supervisor who shoots Victim instead. In that case, $Supervisor=1$ directly NESS-causes $Victim=1$ (also, $Victim=1$ counterfactually depends on $Supervisor=1$). Yet, as the reader may verify, Halpern's definition does not consider $Supervisor=1$ a cause of $Victim=1$.
}

\citet{halpern08} is the latest to offer an attempt at formalizing the NESS definition in the structural equations framework. Unfortunately, just as the attempt discussed below, it was developed before Wright's substantial clarifications in his 2011 article, and therefore it shouldn't come as a surprise that both of them miss the mark. There are two problems with his attempt, only one of which is conclusive evidence that his definition is not a correct formalization of the NESS test. Rather than presenting his definition, I simply state the problems, and leave it to the reader to verify that these apply.

The first problem is only a potential problem, which requires further investigation to be settled: it is not clear whether his definition is transitive. If it is not, then just as with Bochman's definition, it conclusively fails as a formalization of the NESS test. Although I was unable to find any violations of transitivity, Halpern does not mention transitivity and nothing in his formulation suggests that he intended his definition to be transitive.\footnote{In personal communication he has confirmed that he did not intend his definition to be transitive and has not looked into whether it is.} The first problem does not allow us to reach a conclusive verdict, but the second one does:  Halpern's definition violates {\bf Dependence} (and the NESS definition does not, see Section \ref{sec:bv}.) Consider again Example \ref{ex:backup}, but looking at the context in which $Trainee=0$, and thus it is Supervisor who shoots Victim instead. In that case, $Supervisor=1$ directly NESS-causes $Victim=1$ (also, $Victim=1$ counterfactually depends on $Supervisor=1$). Yet, as the reader may verify, Halpern's definition does not consider $Supervisor=1$ a cause of $Victim=1$.

\subsection{Baldwin and Neufeld}

\citet{baldwin} were the first to suggest a formalization of the NESS definition in the structural equations framework. Their attempt faces two problems.

The first problem runs deeper than just the discussion about the NESS definition: their definition makes use of syntactic properties of a structural equation, which is at odds with the manner in which structural equations are usually interpreted \cite[p. 314-315]{pearl:book2}. To illustrate this point, consider the equation $E = A \land B$ and the context in which both $A=0$ and $B=0$. In this case, the NESS definition judges both $A=0$ and $B=0$ to be causes of $E=0$. However, Baldwin and Neufeld's definition judges neither to be causes. But if we write the equation as $\lnot E = \lnot A \lor \lnot B$, then their definition does judge both to be causes. However, both equations are semantically equivalent, and there is no indication in Wright's work that he considers such syntactic properties to be relevant.

The second problem is that their definition is an HP-style definition, because it satisfies all four assumptions discussed in Section \ref{sec:hpstyle}.
As was noted before, this is inconsistent with Wright's views of the NESS definition.

\section{The BV Definition of Causation}\label{sec:bv}

All of Definitions \ref{contribute1}, \ref{nesscause1}, and \ref{nesscause2}, appear explicitly in the work of BV, be it under different names \cite{beckers17,beckers18a}. A direct NESS cause is called ``a direct actual contributor'' and a NESS cause is called ``an actual contributor''. BV do not mention a connection to the NESS definition, for the simple reason that we were unaware such a connection existed.

The BV definition can now be formulated as follows.

\dfn\label{bv1} [BV definition] $C=c$ {\em BV-causes $E=e$ w.r.t.~$(M,\vec{u})$} if $C=c$ NESS-causes $E=e$ w.r.t.~$(M,\vec{u})$ and there exists a $c' \in \R(C)$ such that $C=c'$ does not NESS-cause $E=e$ w.r.t.~$(M_{C \gets c'},\vec{u})$.
\edfn

Several observations are apparent immediately. The first is that being a NESS-cause is a necessary condition for being a BV-cause. \commentout{
BV even go as far as claiming that when we restrict attention to binary variables, this also holds for all of the definitions that we have been referring to as HP-style definitions. Although the claim is true for many examples, the following example shows that it is false in general.

We have as equations $E=(C \land D) \lor A$ and $D=A$. Now consider a context such that $A=1$, $C=1$, and thus also $E=1$. It is easy to see that $C=1$ is a (direct) NESS-cause of $E=1$: $\{C=1,D=1\}$ is sufficient for $E=1$, and $\{D=1\}$ is not. Yet most of the HP-style definitions do not consider $C=1$ to be a cause of $E=1$. 
Recall that {\bf Interventionism} demands $E=1$ counterfactually depends on $C=1$ under some intervention that satisfies certain conditions $P$. It is easy to see that the only possible intervention is $[D \gets 1, A \gets 0]$. The details of the conditions $P$ differ between the definitions, so we focus on the HP definition itself. It demands that  $E=1$ holds under all interventions that include $C=1$ and any subset of $\{D=1,A=0\}$, which is false: we get $E=0$ under the intervention $[C \gets 1, A \gets 0]$.
}
BV even go as far as claiming that when we restrict attention to binary variables, this also holds for all of the definitions that we have been referring to as HP-style definitions. Although the claim is true for many examples, the following example shows that it is false in general.

We have as equations $E=(C \land D) \lor A$ and $A=\lnot D$. Now consider a context such that $D=0$ and $C=1$, and thus also $A=1$ and $E=1$. It is easy to see that $C=1$ is not a NESS-cause of $E=1$: $A=1$ is sufficient by itself, and $\{C=1,D=0\}$ is not sufficient for $E=1$. Yet most of the HP-style definitions do consider $C=1$ to be a cause of $E=1$. Recall that {\bf Interventionism} demands $E=1$ counterfactually depends on $C=1$ under some intervention, and this intervention satisfies certain conditions $P$. The intervention $[D \gets 1]$ clearly meets the first requirement. As details of the conditions $P$ differ between the definitions, we focus on the standard HP definition itself. It demands that $E=1$ holds in this context under all interventions that consist of $C=1$, any subset of the chosen intervention $[D \gets 1]$, and any subset of the intervention that sets the other variables to their actual value, i.e., $E=1$ has to hold in this context for $[C \gets 1]$, $[C \gets 1, D \gets 1]$, $[C \gets 1, A \gets 1]$, $[C \gets 1, D \gets 1, A \gets 1]$. This demand is met.

The following result from \cite{beckers17} allows us to make two more observations.

\begin{theorem}\label{depness} $E=e$ is counterfactually dependent on $C=c$ w.r.t.~$(M,\vec{u})$ iff $C=c$ NESS-causes $E=e$ w.r.t.~$(M,\vec{u})$ and there exist $c' \in \R(C)$, $e' \neq e \in \R(E)$, such that $C=c'$ NESS-causes $E=e'$ w.r.t.~$(M_{C \gets c'},\vec{u})$.
\end{theorem}

This result implies a second observation: both the BV definition and the NESS definition satisfy {\bf Dependence}. Yet neither definitions are HP-style definitions. What makes the NESS definition distinct from HP-style definitions is that it does not satisfy {\bf Counterfactual} or {\bf Interventionism}.

\begin{theorem}\label{thm:nesscause} NESS satisfies {\bf Dependence}, but does not satisfy {\bf Counterfactual} or {\bf Interventionism}.
\end{theorem}

\prf {\bf Dependence} is a direct consequence of Theorem \ref{depness}.

{\bf Counterfactual} and {\bf Interventionism}.
Say we have the following equations: $E =  D \lor \lnot C$ and $D = C$. Consider the context such that $C=1$. Then, according to any definition that accepts {\bf Dependence}, it holds that $C=1$ causes $D=1$ and $D=1$ causes $E=1$. By transitivity, $C=1$ NESS-causes $E=1$. Nevertheless, we also have that $C=0$ is sufficient for $E=1$ in the counterfactual setting $[C \gets 0]$, thus violating {\bf Counterfactual}. Further, since $D$ is the only variable besides $C$ and $E$, the only two interventions that are possible candidates for satisfying {\bf Interventionism} are $D \gets 1$ and $D \gets 0$, neither of which work. 
\eprf

\commentout{
What makes the NESS definition distinct from HP-style definitions is that it does not satisfy {\bf Counterfactual} or {\bf Interventionism}, as can be seen from the following example. 

Say we have the following equations: $E =  D \lor \lnot C$ and $D = C$. Consider the context such that $C=1$. Then, according to any definition that accepts {\bf Dependence}, it holds that $C=1$ causes $D=1$ and $D=1$ causes $E=1$. By transitivity, $C=1$ NESS-causes $E=1$. Nevertheless, we also have that $C=0$ is sufficient for $E=1$ in the counterfactual setting $[C \gets 0]$, thus violating {\bf Counterfactual}. Further, since $D$ is the only variable besides $C$ and $E$, the only two interventions that are possible candidates for satisfying {\bf Interventionism} are $D \gets 1$ and $D \gets 0$, neither of which work. 
}

The BV definition on the other hand is distinct from HP-style definitions only because it does not satisfy {\bf Interventionism}.
\begin{theorem}\label{thm:asymnesscause} BV satisfies {\bf Dependence} and {\bf Counterfactual}, but does not satisfy {\bf Interventionism}.
\end{theorem}

\prf {\bf Dependence}
Assume that $E=e$ is counterfactually dependent on $C=c$ w.r.t.~$(M,\vec{u})$. Then, by Theorem \ref{depness}, we know that $C=c$ NESS-causes $E=e$ w.r.t.~$(M,\vec{u})$ and there exists a $c' \in \R(C)$, $e' \neq e \in \R(E)$, such that $C=c'$ NESS-causes $E=e'$ w.r.t.~$(M_{C \gets c'},\vec{u})$. From the latter it follows that $(M_{C \gets c'},\vec{u}) \sat E = e'$, and thus $C=c'$ does not NESS-cause $E=e$ w.r.t.~$(M_{C \gets c'},\vec{u})$. 

{\bf Counterfactual}
Assume that $C=c$ BV-causes $E=e$ w.r.t.~$(M,\vec{u})$. Say $c' \in \R(C)$ is such that $C=c'$ does not NESS-cause $E=e$ w.r.t.~$(M_{C \gets c'},\vec{u})$. We proceed by a reductio: assume that $C=c'$ is sufficient for $E=e$ w.r.t.~$(M_{C \gets c'},\vec{u})$. By Definition \ref{nesscause1}, this means that either $C=c'$ is a direct NESS-cause of $E=e$ w.r.t.~$(M_{C \gets c'},\vec{u})$, or that $\emptyset$ is sufficient for $E=e$ w.r.t.~$(M_{C \gets c'},\vec{u})$. The former contradicts our assumption, so it has to be the latter. In other words, $f_{E}(\vec{u})=e$. But this implies that $E=e$ has no direct NESS causes, and thus also no NESS-causes. This contradicts the assumption that $C=c$ BV-causes $E=e$.

{\bf Interventionism}
Assume we have a model with the following equations (all variables are binary): $E = F \lor \lnot A \land \lnot D$, $F=D$, and $D=C \lor A$. Say the context is such that $A=1$ and $C=1$ hold. Then clearly $C=1$ BV-causes $E=1$. Yet there is no intervention such that $E=1$ counterfactually depends on $C=1$ given that intervention. To see why, note that such an intervention could not include $D$ or $F$, for any influence of $C$ on $E$ goes through those variables. So the only candidates are $A \gets 1$ and $A \gets 0$, neither of which work.
\eprf

A third observation is that both the BV definition and counterfactual dependence consist of the NESS definition combined with a counterfactual difference-making condition. Concretely, they both require that there exists a counterfactual value which fulfills a different role than the actual value. In the case of counterfactual dependence, that different role is to NESS-cause a counterfactual value of the effect. The BV definition merely demands that the counterfactual value does not fulfill the same role as the actual value, without further specification. Thus the BV definition represents a compromise between counterfactual dependence and NESS-causation: it adds a counterfactual component to NESS-causation that guarantees causes make a difference to their effect, but does not demand that this difference consists in preventing the effect from occurring. 

\section{The CNESS Definition of Causation}\label{sec:problems}

Consider again Example \ref{ex:backup}. We already concluded that $Trainee=1$ NESS-causes $Victim=1$. To assess whether $Trainee=1$ also BV-causes $Victim=1$, we have to look at the counterfactual scenario that is the result of the intervention $[Trainee \gets 0]$. In that scenario, $Trainee=0$ directly NESS-causes $Supervisor=1$, which directly NESS-causes $Victim=1$, leading to the conclusion that $Trainee=0$ also NESS-causes $Victim=1$. So Victim dies either way. More important for our purposes (but probably not for Victim) is that the candidate cause does not meet BV's difference-making condition, and thus $Trainee=1$ does not BV-cause $Victim=1$. 

BV show in detail that the same verdict is reached for other examples of early preemption \cite{beckers17}. Although BV attempt to justify this verdict by appealing to probabilistic considerations, there is a cheaper -- and arguably more compelling -- solution: adopt a more subtle counterfactual difference-making condition. The idea is to capture the role of some event $C=c$ in a more nuanced way than merely stating that it NESS-causes an effect $E=e$, by looking at the specific path along which the chain of direct NESS causes proceeds. 

\dfn\label{nesscause3} {\em $C=c$ NESS-causes $E=e$ along a path $p$} w.r.t.~$(M,\vec{u})$ if the values of the variables in $p$ form a chain of direct NESS causes from $C=c$ to $E=e$. (I.e., there exists a path $p=(C_1,\ldots, C_n)$, and values $c_1,\ldots,c_n$, so that $C=c$ is a direct NESS cause of $C_1=c_1$, $\ldots$, and $C_n=c_n$ is a direct NESS cause of $E=e$.)
\edfn

By focussing on a specific path and its subpaths, it becomes possible to formulate a more subtle difference-making condition. Say $C=c$ NESS-causes $E=e$ along a path $p$, and there is some $C=c'$ that would also NESS-cause $E=e$ along some path $q$ (and not along any other path). If $q$ contains variables that do not appear in $p$, then $C=c$ and $C=c'$ do not fulfill the same causal role, and thus $C=c$ made a difference to the manner that the effect $E=e$ came about. Implementing this idea results in the {\em Counterfactual NESS definition}, which I suggest as an improvement of the BV definition.

\dfn\label{cness} [CNESS definition] $C=c$ {\em CNESS-causes $E=e$ w.r.t.~$(M,\vec{u})$} if $C=c$ NESS-causes $E=e$ along some path $p$ w.r.t.~$(M,\vec{u})$ and there exists a $c' \in \R(C)$ such that $C=c'$ does not NESS-cause $E=e$ along any subpath $p'$ of $p$ w.r.t.~$(M_{C \gets c'},\vec{u})$.  (A subpath of $p$ is a path whose variables are all members of $p$.)
\edfn

\commentout{
It is easy to verify that the proof of Theorem \ref{thm:asymnesscause} can be used without change to show that the CNESS definition also satisfies {\bf Dependence} and {\bf Counterfactual}, but does not satisfy {\bf Interventionism}.
}

\begin{theorem}\label{thm:asymnesscause2} CNESS satisfies {\bf Dependence} and {\bf Counterfactual}, but does not satisfy {\bf Interventionism}.
\end{theorem}

\prf
The proof of Theorem \ref{thm:asymnesscause} applies without change. (Note that direct NESS-causes are NESS-causes along the empty path.)
\eprf

Applying this definition to Example \ref{ex:backup}, we see that $Trainee=1$ NESS-causes $Victim=1$ directly (i.e., along the empty path), whereas $Trainee=0$ NESS-causes $Victim=0$ along $Supervisor$. As a result, $Trainee=1$ CNESS-causes $Victim=1$. 

Lastly, although I have refrained from discussing the temporal considerations that BV invoke to deal with cases of Late Preemption, I point out that the CNESS definition handles standard cases of Late Preemption as the following one without invoking such temporal considerations, suggesting that it also offers an alternative to the BV definition-2 \cite{beckers18a}.

\begin{example}\label{LP}
Suzy and Billy both throw a rock at a bottle. Suzy's rock gets there first, shattering the bottle. However Billy's throw was also accurate, and would have shattered the bottle had it not been preempted by Suzy's throw.
\end{example}

\cite{halpernpearl05a} use the following variables for this example, which capture the fact that Billy's throw was preempted by Suzy's rock hitting the bottle: $BS$ for the bottle shattering, $BH$, $SH$ for Billy's (resp. Suzy's) rock hitting the bottle, and two more variables ($BT$, $ST$) for either of them throwing their rock. The equations are then as follows: $BS=BH \lor SH$, $SH=ST$, $BH=BT \land \lnot SH$. The reader may verify that in the context under consideration in which $ST=1$ and $BT=1$, the BV definition offers the mistaken verdict that $ST=1$ does not cause $BS=1$, whereas the CNESS definition does not. (Note also that in the scenario where Suzy {\em does not throw}, the NESS definition judges $ST=0$ to be a cause of $BS=1$. This is an unacceptable result that the BV and CNESS definitions avoid.)

\section{Conclusion}

I have formalized Wright's NESS definition of causation using the framework of structural equations modeling, thereby showing that this framework is not exclusively suitable for counterfactual approaches. Moreover, this formalization revealed that the recent approach by Beckers \& Vennekens offers a nice compromise between a regularity approach and the most popular counterfactual approaches based on that of Halpern \& Pearl. Further spelling out the relations between all these definitions allowed a modification of the BV definition that removes a major disadvantage of that approach. The resulting Counterfactual NESS definition of causation combines the NESS definition with a subtle counterfactual difference-making condition. In future work I intend to explore further properties of this definition.

\subsubsection{Acknowledgments.}

Many thanks to Hein Duijf, Joe Halpern, Joost Vennekens, and Marc Denecker for comments on earlier versions of this paper. This research was made possible by funding from the Alexander von Humboldt Foundation.

\bibliographystyle{spbasic} 
\bibliography{joe,allpapers}

\end{document}